\documentclass[letterpaper, 10 pt, conference]{ieeeconf} 

\IEEEoverridecommandlockouts 

\usepackage{graphicx} 
\graphicspath{{./figures/}}
\usepackage{mathptmx} 
\usepackage{amsmath} 
\usepackage{amssymb} 
\usepackage[skip=0.3ex]{subcaption}
\usepackage{placeins}
\usepackage{float}
\usepackage{xcolor}
\usepackage{lipsum}

\DeclareMathAlphabet      {\mbi}{OML}{cmm}{b}{it}
\DeclareMathAlphabet      {\mathit}{OML}{cmm}{}{it}
\DeclareMathAlphabet{\mathcal}{OMS}{cmsy}{m}{n}
\DeclareMathOperator{\Dh}{{}^\mathcal{\Tilde{D}}}
\DeclareMathOperator{\D}{{}^\mathcal{D}}
\DeclareMathOperator{\AdE}{\text{\textit{\textbf{Ad}}}_{\mbi{g}_E}}
\DeclareMathOperator{\expO}{{\text{exp}}_{SO(3)}}
\DeclareMathOperator{\expE}{{\text{exp}}_{SE(3)}}

\newcommand\numberthis{\addtocounter{equation}{1}\tag{\theequation}}

\begin{document}
%
\title{\LARGE \bf Multi-DoF Time Domain Passivity Approach Based Drift Compensation for Telemanipulation}
%
%
%

\author{Andre Coelho$^1$, Christian Ott$^1$, Harsimran Singh$^1$,  Fernando Lizarralde$^2$, Konstantin Kondak$^1$
\thanks{$^{1}$The authors are with the Institute of Robotics and Mechatronics of the
German Aerospace Center (DLR), Oberpfaffenhofen, Germany}%
\thanks{$^{2}$The author is with the Department of Electrical Engineering,
COPPE, Federal University of Rio de Janeiro (UFRJ), Brazil}
\thanks{{\tt\small Andre.Coelho@dlr.de}}}

\maketitle

\begin{abstract}
When, in addition to passivity, position synchronization is also desired in bilateral teleoperation, Time Domain Passivity Approach (TDPA) alone might not be able to fulfill the desired objective. This is due to an undesired effect caused by admittance type passivity controllers, namely position drift. Previous works focused on developing TDPA-based drift compensation methods to solve this issue. It was shown that, in addition to reducing drift, one of the proposed methods was able to keep the force signals within their normal range, guaranteeing the safety of the task. However, no multi-DoF treatment of those approaches has been addressed. In that scope, this paper focuses on providing an extension of previous TDPA-based approaches to multi-DoF Cartesian-space teleoperation. An analysis of the convergence properties of the presented method is also provided. In addition, its applicability to multi-DoF devices is shown through hardware experiments and numerical simulation with round-trip time delays up to 700 ms. 
\end{abstract}

\IEEEpeerreviewmaketitle

\setlength{\textfloatsep}{10pt}
\section{Introduction}
Despite being able to autonomously fulfill a significant range of objectives, state-of-the-art robots still need human assistance to perform more complex or unforeseen tasks \cite{schmaus18}. The level of human participation in robotic tasks can range from supervised autonomy \cite{lii17} to direct teleoperation \cite{artigas16}. In the latter, an important characteristic of the telemanipulation setup is to be able to passively interact with the environment and the human operator. Among the passivity-based telemanipulation approaches (e.g. \cite{anderson89,niemeyer91}) developed to solve that issue, Time Domain Passivity Approach (TDPA, \cite{hannaford02,ryu10}) presents the advantage of adapting the energy dissipation necessary to passivate the teleoperation channel based on measurements of the flow and effort variables acting on the system. This characteristic allows the implementation of a model-independent passivity observer and passivity controller (PO-PC) pair, which is robust to varying time delays and package loss in the communication channel. The adaptive characteristic of TDPA results in better performance compared to other passivity-enforcing controllers for teleoperation, e.g. wave-variable methods (see  \cite{balachandran16}).
\par
Nevertheless, in spite of being able to render the communication channel passive, TDPA presents the drawback of creating position drift between master and slave devices whenever the PO-PC pair is applied in admittance configuration, which is necessary in many telemanipulation architectures (\cite{artigas16,ryu10, artigas10pp}). In order to tackle this issue, Artigas et al. \cite{artigas10} proposed a modification to the traditional TDPA approach in order to inject energy into the system to compensate for the existing drift. Later, Chawda et al. \cite{chawda14} adapted Artigas' compensator in order keep the original TDPA formulation by using a virtual velocity injection source before the PO-PC. Despite being able to successfully compensate for the drift, these methods generate force spikes when the compensation action is allowed into the system after drift has been accumulated. In order to achieve position synchronization while keeping the forces within their normal range, a TDPA-based drift compensator was developed by the authors \cite{coelho18}. In that paper, the previously existing compensation methods were adapted to produce smoother signals while removing the drift. The efficacy of TDPA-based compensation methods was experimentally shown through application to one-degree-of-freedom (1-DoF) devices or in a concatenated manner, treating each DoF as an independent system. Nevertheless, no multi-DoF application of those compensators has been tackled to this date. 
\par
\begin{figure}[t]
\centering
\includegraphics[trim={0cm 0cm 0cm 0cm},clip,width=0.9\linewidth]{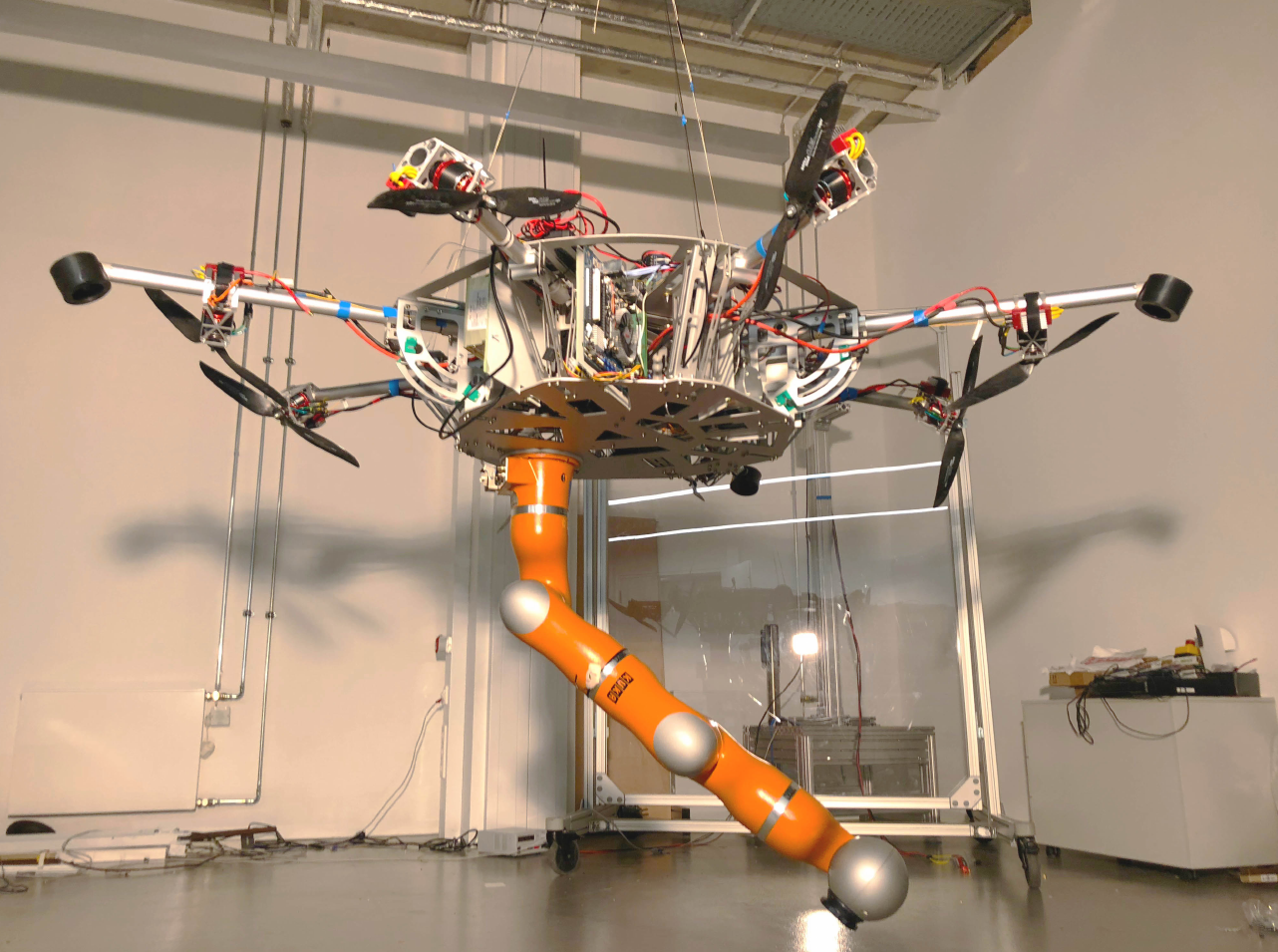}
\caption{DLR Suspended Aerial Manipulator \cite{sarkisov2019}, whose model was used to validate the proposed approach.}
\label{fig:sam}
\end{figure}

In light of that, this paper aims at providing an extension of the previously presented drift compensators \cite{artigas10,chawda14, coelho18} to multi-DoF robotic systems. In addition, a convergence analysis is provided. It is shown that, if the gains are kept within a given range and if allowed by the passivity condition, the presented method is able to successfully reduce the accumulated drift caused by admittance type passivity controllers (PCs) in TDPA. In addition to hardware experiments with commercially available \textit{Novint Falcon} haptic devices, teleoperation of the dynamic model of a Suspended Aerial Manipulator \cite{sarkisov2019} (see Fig.~\ref{fig:sam}) is simulated.
\par
Together with the previously presented single-DoF analyses, this paper contributes to demonstrating the effectiveness of the TDPA-based drift compensators and their applicability to different teleoperation setups. 

\section{Dynamics of Kinematically Redundant Manipulators}
\label{sec:dyn}
The dynamic model of a robotic manipulator with $n$ joints can be written as
\begin{equation}
   \mbi{ M(q)\Ddot{q}+C(q,\dot{q})\dot{q}+g(q)=\tau} \, ,
\end{equation}
where $\mbi{q} \in \mathbb{R}^n$ is a set of generalized coordinates, $\mbi{M(q)} \in \mathbb{R}^{n \times n}$ is the inertia matrix, $\mbi{C(q,\dot{q})\dot{q}} \in \mathbb{R}^n$ is a vector of Coriolis and centrifugal forces, and $\mbi{g(q)} \in \mathbb{R}^n$ is the gravitational generalized torque vector. The generalized torque vector $\mbi{\tau} \in \mathbb{R}^n$ is the sum of control and external torques.
\par
In case kinematically redundant robots are considered, where the minimum number of local Cartesian task coordinates $m$ is less than the number of joint generalized coordinates $n$, the set of task coordinates can be defined as
\begin{equation}
    \begin{bmatrix}
    \mbi{v_x} \\
    \mbi{v_n} \\
    \end{bmatrix}
    =
    \mbi{\Bar{J}(q)\dot{q}}=
    \begin{bmatrix}
    \mbi{J(q)} \\
    \mbi{N(q)} \\
    \end{bmatrix} \mbi{\dot{q}} \, ,
\end{equation}
where $\mbi{v_x} \in \mathbb{R}^{m}$ and $\mbi{v_n} \in \mathbb{R}^{n-m}$ represent the Cartesian and null space velocities, respectively. $\mbi{J(q)}$ is the Jacobian matrix that maps generalized to Cartesian coordinates. Under the assumption of full row rank of $\mbi{J(q)}$, the matrix $\mbi{N(q)}$ can be constructed as \cite{ott15}
\begin{equation}
\label{eq:N}
    \mbi{N(q)=(Z(q)M(q)Z(q)}^\mathit{T})^{-1}\mbi{Z(q)M(q)} \, ,
\end{equation}
where $\mbi{Z(q)}$ is a full row rank nullspace base matrix, such that $\mbi{J(q)Z(q)}^\mathit{T}=\mathbf{0}$. 
\par
Such formulation allows the manipulator dynamics to be written as 
\begin{align*}
    & \begin{bmatrix}\mbi{\Lambda_x(q)} & 0 \\ 0 & \mbi{\Lambda_n(q)} \\ \end{bmatrix} \begin{bmatrix}\mbi{\dot{v}_x} \\ \mbi{\dot{v}_n} \\\end{bmatrix} + \begin{bmatrix}\mbi{\mu_x(q,\dot{q})} & \mbi{\mu_{xn}(q,\dot{q})} \\ \mbi{\mu_{nx}(q,\dot{q})} & \mbi{\mu_n(q,\dot{q})} \\ \end{bmatrix}\begin{bmatrix}\mbi{v_x} \\ \mbi{v_n} \\\end{bmatrix} \\ &= \mbi{\Bar{J}(q)}^\mathit{-T}\mbi{(\tau-g(q))}\, .  \numberthis
\end{align*}
\par
The choice of $\mbi{N(q)}$ as in (\ref{eq:N}) generates a block diagonal matrix $\mbi{\Lambda(q)}$. By compensating the gravity torque $\mbi{g(q)}$ and the cross-coupling terms of $\mbi{\mu(q,\dot{q})}$ such that the nullspace task has no influence on the Cartesian one \cite{ott15}, the dynamics of the Cartesian-space task can be rewritten as 
\begin{equation}
\label{eq:cart_dyn}
    \mbi{\Lambda_x(q)\dot{v}_x+\mu_x(q,\dot{q})v_x  = F_x}\, ,
\end{equation}
where $\mbi{F_x} \in \mathbb{R}^m$ is a Cartesian-space wrench.
\par
The above described dynamic decoupling allows for the application of TDPA for Cartesian-space teleoperation of kinematically redundant manipulators without having to take the energy generated by the nullspace task into account.

\section{Time Domain Passivity Approach}
\label{sec:tdpa}
\subsection{Overview}
In contrast to methods where a damping element is designed for the worst case scenario (\cite{anderson89,niemeyer91}), TDPA consists in adding adaptive damping components in order to dissipate only the necessary amount of energy, computed using measurements of the forces and velocities being exchanged.
\par
\begin{figure}[t]
\centering
\includegraphics[trim={7.4cm 8cm 9.6cm 7.4cm},clip,width=0.9\linewidth]{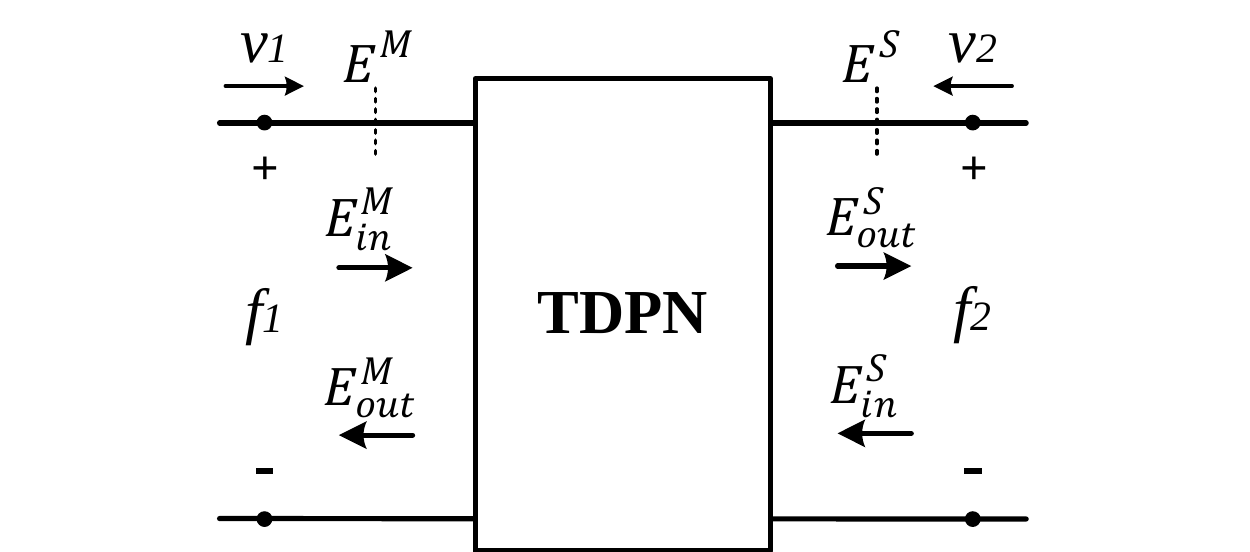}
\caption{Signal flow of the TDPN.}
\label{fig:tdpn_flow}
\end{figure}

In TDPA, the communication channel is usually represented by one or more Time Delay Power Networks (TDPNs, \cite{ryu10}), which are two port networks that exchange velocities and forces. In addition to constant or variable time delays, TDPNs can also model package losses in the signals being transmitted. Fig.~\ref{fig:tdpn_flow} shows the signal flow of the TDPN. $E^M$ and $E^S$ are the energies computed on the master and slave sides, respectively. The $in$ and $out$ subscripts are used to represent the direction of flow, namely into or out of the channel. 
\par
 The pairs $v_1$/$f_1$ and $v_2$/$f_2$ from Fig.~\ref{fig:tdpn_flow} are the flow-effort pairs on each side of the TDPN, such that
 \begin{align} \label{eq:energy_integrals}
E^M(k)  &= \Delta T \sum_{j=0}^{k} f_1(j)^\mathit{T} v_1(j)\, , \\
E^S(k) &= -\Delta T \sum_{j=0}^{k} f_2(j)^\mathit{T} v_2(j)\, ,
\end{align}
where $\Delta T$ is the sampling time.
\par
A sufficient condition for passivity of a TDPN network is that
\begin{align} 
E^{L2R}_{obs}(k) = E^M_{in}(k-T_f(k))-E^S_{out}(k) \geq 0, \quad \forall k \geq 0, \label{eq:energy_obs} \\
E^{R2L}_{obs}(k) = E^S_{in}(k-T_b(k))-E^M_{out}(k) \geq 0, \quad  \forall k \geq 0, \label{eq:energy_obs1}
\end{align} 
where $E^{L2R}_{obs}(k)$ and $E^{R2L}_{obs}(k)$ are the observed left-to-right and right-to-left energy flows observed on the right and left-hand sides of the TDPN. $T_f$ and $T_b$ are the forward and backward delays, respectively.
\par
\begin{figure}[tbhp]
\centering
\includegraphics[trim={6cm 6.7cm 4.8cm 6cm},clip,width=1\linewidth]{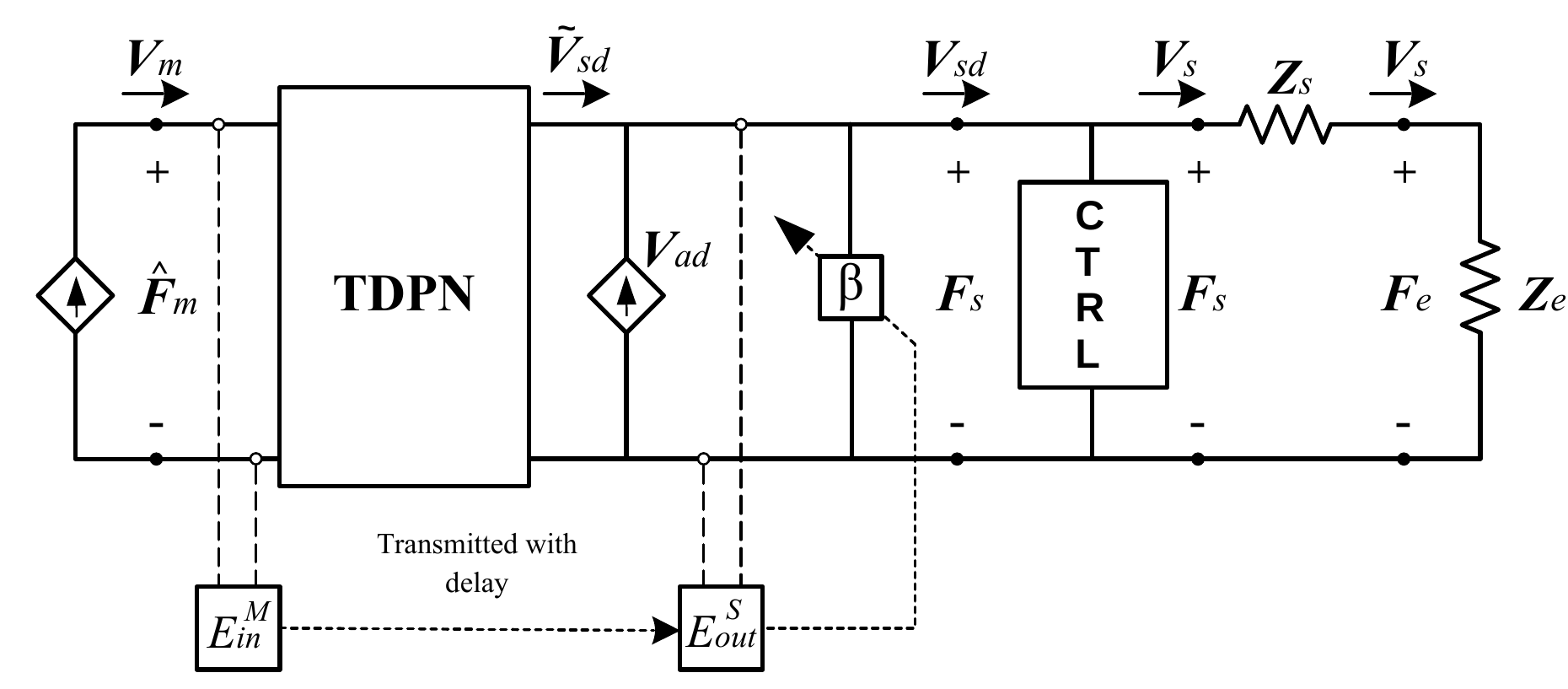}
\caption{Slave side of a P-F architecture. The PO-PC pair ($\mbi{\beta}$) is applied in admittance configuration. $\mbi{V}_{ad}$ is the drift compensator velocity.}
\label{fig:drift_comp}
\end{figure}

One of the most common teleoperation schemes is the P-F architecture \cite{lawrence93}, where the master velocity is sent through the channel and serves as desired velocity to the slave. In turn, the force produced by the slave-side controller is sent back to the master. Following the framework presented by Artigas et al. \cite{artigas11}, using a hybrid of circuit and network representation, the slave side of the P-F architecture can be represented as shown in Fig.~\ref{fig:drift_comp}. There, the communication channel is represented by a TDPN. $\mbi{V}_m$ and $\mbi{V}_s$ are the velocities of the master and slave devices. $\mbi{F}_s$ is the force exerted by the slave-side controller and $\hat{\mbi{F}}_m$ is its delayed version applied to the master device. $\mbi{\beta}$ and $\mbi{V}_{ad}$ are the admittance-type passivity controller and the drift compensation velocity source, which will be addressed subsequently. $\mbi{\Tilde{V}}_{sd}$ is the delayed master velocity and $\mbi{V}_{sd}$ is the velocity given as a reference to the slave controller after being modified by the drift compensator and the passivity controller.

\subsection{Passivity Observer}
In order to take into account the energy removed by the passivity controllers up to the previous time steps ($E^M_{PC}(k-1)$ and $E^S_{PC}(k-1)$), the energy flow on each side of the TDPN is computed as 
\begin{align} \label{eq:po}
W_M(k) = E^S_{in}(k-T_b(k))-E^M_{out}(k)+E^M_{PC}(k-1), \\
W_S(k) = E^M_{in}(k-T_f(k))-E^S_{out}(k)+E^S_{PC}(k-1). 
\label{eq:po2}
\end{align} 
\subsection{Passivity Controller}
\label{sec:pc}
The passivity controller acts as an adaptive damping in order to guarantee the passivity of the channel. It can be applied in impedance or admittance configuration, according to the architecture requirements. In Fig.~\ref{fig:drift_comp} the PC ($\mbi{\beta}$) is being applied in admittance configuration in order to modify the velocity coming out of the channel. 

\par
In order to fulfill the passivity conditions from (\ref{eq:energy_obs}) and (\ref{eq:energy_obs1}) for the Cartesian-space task, two constructions for the passivity controller are possible, namely the concatenated version and the coupled one.
\subsubsection{Concatenated PO-PC}
The concatenated version consists of adding a passivity observer to each degree of freedom $(W^i_S(k))$ and computing $\mbi{\beta}$ as a diagonal matrix, whose diagonal elements $\beta^i$ are given by
\begin{equation} \label{eq:beta}
\beta^i(k)=
\begin{cases}
    0                                                        & \text{if } W^i_S(k)>0\\
    -\cfrac{ W^i_S(k)}{\Delta TF^i_s(k)^2} & \text{else, if } |F^i_s(k)|>0 \,,
\end{cases}
\end{equation}
where $\Delta T$ is the sampling time.
\subsubsection{Coupled PO-PC}
In addition to the concatenated version, the passivity controller can be applied in a coupled manner. For that purpose, the impedance PC presented by Ott et al. \cite{ott11} can be adapted to the admittance case as
\begin{equation}
    \mbi{\beta}(k)=d_f(k)\mbi{\Lambda}_x(\mbi{q}(k))^{-1} \, ,
\end{equation}
\begin{equation} \label{eq:beta}
d_f(k)=
\begin{cases}
    0                                                        & \text{if } W_S(k)>0\\
    -\cfrac{ W_S(k)}{\Delta T||\mbi{F}_{s}(k)||^2_f} & \text{else, if } ||\mbi{F}_{s}(k)||^2_f>0 \,,
\end{cases}
\end{equation}
where
\begin{equation}
    ||\mbi{F}_{s}(k)||^2_f=\mbi{F}_{s}(k)^\mathit{T}\mbi{\Lambda}_x(\mbi{q}(k))^{-1}\mbi{F}_{s}(k) \, .
\end{equation}
\par
For both cases, the velocity removed by the passivity controller from the delayed master velocity in order to keep the system passive will be
\begin{equation}
    \mbi{V}_{pc}(k)=\mbi{\beta}(k)\mbi{F}_{s}(k) \, ,
\end{equation}
and the resulting velocity used as a reference by the slave will be
\begin{equation} \label{eq:pc_adm}
\mbi{V}_{sd}(k)=\Tilde{\mbi{V}}_{sd}(k)- \mbi{V}_{pc}(k)\, ,
\end{equation}
assuming all velocities are represented in the same frame. 

\section{Multi-DoF Drift Compensator}
\label{sec:comp}
\subsection{Notations and Definitions}
\label{sec:notation}
\subsubsection{The Special Euclidean group and its Lie algebra} The pose of a rigid body in space can be represented by the special Euclidean Lie group $SE(3)$, whose elements are of the form
\begin{equation}
    \mbi{g}=\begin{bmatrix}\mbi{R} &  \mbi{p} \\\mathbf{0} & 1\\\end{bmatrix} \in \: SE(3) \, ,
\end{equation}
where $\mbi{p}$ is a vector in $\mathbb{R}^3$ and $\mbi{R}$ is an element of the Special Orthogonal group $SO(3)$, whose Lie algebra is $\mathfrak{so}(3)$.
Furthermore, the velocity of a rigid body can be expressed by elements of the Lie algebra of $SE(3)$, namely $\mathfrak{se}(3)$, as
\begin{equation}
   [\mbi{V}]^\wedge=\begin{bmatrix}\widehat{\mbi{\omega}} &  \mbi{v} \\\mathbf{0} & 0\\\end{bmatrix} \: \in \mathfrak{se}(3) \, ,
\end{equation}
where $ \widehat{\cdot}$ indicates the skew-symmetric operator applied to a vector in $\mathbb{R}^3$, and $\mbi{\omega}, \, \mbi{v} \in \mathbb{R}^3$ are angular and linear velocities, respectively. Adding to that, due to the isomorphism between $\mathfrak{se}(3)$ and $\mathbb{R}^6$, it is useful to define the operators $[\cdot]^\wedge \, : \, \mathbb{R}^6 \rightarrow \mathfrak{se}(3)$ and $[\cdot]^\vee \, : \,  \mathfrak{se}(3) \rightarrow \mathbb{R}^6$, such that the velocity of a rigid body can be expressed as $ \mbi{V}=\begin{bmatrix} \mbi{\omega}^\mathit{T} &\mbi{v}^\mathit{T}\\ \end{bmatrix}^\mathit{T} \in \mathbb{R}^6$, which can be represented in body (${}^B \mbi{V}$) or in spatial frame (${}^S \mbi{V}$) \cite{murray94}. 
\subsubsection{Exponential map} Given $\widehat{\varphi} \in \mathfrak{so}(3)$ and $\mathbf{X}=(\widehat{\varphi},q) \in \mathfrak{se}(3)$, the exponential maps in $SO(3)$ and $SE(3)$ can be defined as \cite{bullo95}
\begin{align}
    \expO (\widehat{\varphi})  = & \mathbf{I}+sin||\varphi||\cfrac{\widehat{\varphi}}{||\varphi||}+\left(1-cos||\varphi||\right)\cfrac{\widehat{\varphi}^2}{||\varphi||^2} \, , \\
    \expE (\mathbf{X})= &\begin{bmatrix}\expO (\widehat{\varphi}) & \mbi{A}(\varphi)q\\ \mathbf{0} & 1\end{bmatrix}\, ,
    \label{eq:exp_se3}
\end{align}
where
\begin{align}
    &\mbi{A}(\varphi) = \mathbf{I}+\left(\cfrac{1-cos||\varphi||}{||\varphi||}\right)\cfrac{\widehat{\varphi}}{||\varphi||}+\left(1-\cfrac{sin||\varphi||}{||\varphi||}\right)\cfrac{\widehat{\varphi}^2}{||\varphi||^2} \, , \\
    &\mbi{A}(\mathbf{0}) = \mathbf{I} \, , \\
    &\mbi{A}(\varphi)^{-1} =  \mathbf{I}-\cfrac{1}{2}\, \widehat{\varphi}+\left(1-\alpha(||\varphi||)\right)\cfrac{\widehat{\varphi}^2}{||\varphi||^2} \, ,
\end{align}
where 
\begin{equation}
\alpha(||\varphi||)\triangleq \frac{||\varphi||}{2}\,\text{cot}\left(\frac{||\varphi||}{2}\right) \, .
\end{equation}
 In addition, a useful identity is 
\begin{equation}
\label{eq:A_id}
    \mbi{A}(\varphi)^{-\mathit{T}} = \mbi{A}(\varphi)^{-1}\expO(\widehat{\varphi})  \, ,
\end{equation}
where
\begin{equation}
\label{eq:A_mT}
    \mbi{A}(\varphi)^{-\mathit{T}} = \mathbf{I}+\cfrac{1}{2}\, \widehat{\varphi}+\left(1-\alpha(||\varphi||)\right)\cfrac{\widehat{\varphi}^2}{||\varphi||^2}  \: .
\end{equation}
\subsubsection{Logarithmic map in $SO(3)$} The logarithmic map of a matrix $\mbi{R} \in SO(3)$ such that $\text{tr}(\mbi{R})\neq -1$ can be defined as
\begin{equation}
    \text{log}_{SO(3)}(\mbi{R})=\cfrac{\gamma}{2sin\gamma}\left( \mbi{R}-\mbi{R}^{\mathit{T}}\right)\:  \in \mathfrak{so}(3) \, ,
\end{equation}
where $cos\gamma=\frac{1}{2}(\text{tr}(\mbi{R})-1)$ and $|\gamma|<\pi$.
\subsubsection{Dynamical systems in $SE(3)$}
A dynamical system with state $\mbi{g} \in SE(3)$ evolves according to the following differential equation in continuous time \cite{bullo95}
\begin{equation}
\label{eq:g_dot}
    \dot{\mbi{g}}(t)=[{}^S\mbi{V}(t)]^\wedge\,\mbi{g}(t)=\mbi{g}(t)\,[{}^B\mbi{V}(t)]^\wedge\, ,
\end{equation}
whose recursive solution in discrete time, given a set of initial conditions, can be approximated to
\begin{align}
\label{eq:int_space}
    \mbi{g}(k)&=\expE\left([{}^S\mbi{V}(k)]^{\wedge}\Delta T\right)\,\mbi{g}(k-1) \, ,\\
    \mbi{g}(k)&=\mbi{g}(k-1)\expE\left([{}^B\mbi{V}(k)]^{\wedge}\Delta T\right) \, .
\label{eq:int_body}
\end{align}
\subsection{Representation of Drift in $SE(3)$}
\label{sec:drift}
Assuming the teleoperation task comprises the complete Cartesian space, the velocities $\Tilde{\mbi{V}}_{sd}(k)$ and $\mbi{V}_{sd}(k)$ can be defined to be \textit{body velocities} \cite{murray94} in~$\mathbb{R}^6$ as 
\begin{equation}
    \Dh \mbi{\Tilde{V}}_{sd}(k)=\begin{bmatrix}\mbi{\omega}_{\Tilde{\mathcal{D}}}\\\mbi{v}_{\Tilde{\mathcal{D}}}\\\end{bmatrix} \, , \quad \D\mbi{V}_{sd}(k)=\begin{bmatrix}\mbi{\omega}_{\mathcal{D}}\\\mbi{v}_{\mathcal{D}}\\\end{bmatrix} \, ,
\end{equation}
where $\mathcal{\Tilde{D}}$ and $\mathcal{D}$ are the frames defined by the delayed master orientation and the orientation given to the slave as the reference, respectively. The discrete-time integral of $\Dh \mbi{\Tilde{V}}_{sd}$ and $\D\mbi{V}_{sd}$ can be computed following (\ref{eq:int_body}) as
\begin{align}
\label{eq:dint}
    \mbi{g}_{\mathcal{D}}(k)=\mbi{g}_{\mathcal{D}}(k-1) \expE\bigl([\D\mbi{V}_{sd}(k)]^{\wedge}\Delta T\bigr) \, , \\
    \mbi{g}_{\mathcal{\Tilde{D}}}(k)=\mbi{g}_{\mathcal{\Tilde{D}}}(k-1) \expE\bigl([\Dh\mbi{\Tilde{V}}_{sd}(k)]^\wedge\Delta T\bigr) \, , 
     \label{eq:dint2}
\end{align}
\par
Using the definitions above, the drift present in the system at a given time step ($k$) can be represented in $SE(3)$ by
\begin{equation}
\label{eq:drift}
    \mbi{g}_{E}(k)=\mbi{g}_{\Tilde{\mathcal{D}}}(k)^{-1}\mbi{g}_{\mathcal{D}}(k)=\begin{bmatrix}  \mbi{R}_E(k) &  \mbi{p}_E(k) \\ \mathbf{0} & 1  \end{bmatrix} \, .
\end{equation}
\par
It can be noted from (\ref{eq:pc_adm}) and (\ref{eq:dint})--(\ref{eq:drift}) that, if the PC acts at a time step, it will affect the value of $\mbi{g}_{E}$ for all future time steps. In case $\mbi{g}_{E}$ is not the identity matrix, there will be a drift between the delayed master pose and the pose given as reference to the slave.


\subsection{Cartesian-Space Drift Compensation}
In order to compensate for the drift caused by TDPA, an additional velocity signal $\mbi{V}_{ad}$ can be added to the delayed master velocity before it is checked by the PO. In Fig.~\ref{fig:drift_comp} the drift compensator is represented by a current source. It can be noted that, since $\mbi{V}_{ad}$ is applied before the point where the energies are computed, the modified velocity $\Tilde{\mbi{V}}_{sd}(k)+\mbi{V}_{ad}(k)$ will be checked and corrected for passivity. This guarantees that the compensation action will only be applied when so-called ``passivity gaps'' appear, i.e., when $W_S(k)$ from (\ref{eq:po2}) would be greater than zero. Therefore, the compensator would not compromise the passivity of the system. From Fig.~\ref{fig:drift_comp}, it can been seen that, when the drift compensator is added, (\ref{eq:pc_adm}) becomes
\begin{equation} \label{eq:drift_added}
\AdE(k) \, \D\mbi{V}_{sd}(k)=  \, \Dh\Tilde{\mbi{V}}_{sd}(k) + \Dh\mbi{V}_{ad} (k) - \Dh\mbi{V}_{pc}(k)\, ,
\end{equation}
where $\AdE(k)$ is defined as \cite{bullo95}
\begin{equation}
    \AdE(k) = \begin{bmatrix} \mbi{R}_E(k) & \mathbf{0}   \\ 
    \widehat{\mbi{p}}_E(k)\mbi{R}_E(k) &  \mbi{R}_E(k) \end{bmatrix} \, .
\end{equation}
\par
In order to reduce the drift between master and slave devices whenever allowed by the aforementioned passivity conditions, the following law can be used
\begin{align} \label{eq:vad}
    \begin{split}
        \mbi{\omega}_{ad}(k)&=-\cfrac{1}{\Delta T}\, \mbi{k_R} \,\varphi_E(k-1) \, , \\ 
         \mbi{v}_{ad}(k)&= -\cfrac{1}{\Delta T} \,  \mbi{A}\left(\omega_{ad}(k) \Delta T\right)^{-\mathit{T}} \mbi{K_T} \, \mbi{p}_E(k-1)\, ,
    \end{split} 
\end{align}
\begin{equation} \label{eq:vad2}
    \Dh\mbi{V}_{ad}(k)=\begin{bmatrix}  \mbi{\omega}_{ad}(k) \\     \mbi{v}_{ad}(k) \end{bmatrix} \, ,
\end{equation}
where $\widehat{\varphi}_E(k-1)=\text{log}_{SO(3)}\left(\mbi{R}_E(k-1)\right)$,
and $\mbi{K_T} \in \mathbb{R}^{3 \times 3}$ and $\mbi{k_R} \in \mathbb{R}$ are the translational and rotational gains of the compensator.  Moreover, $\mbi{A}^{-\mathit{T}}$ is defined in Section~\ref{sec:notation}.

\subsection{Convergence Analysis}

As mentioned in Section~\ref{sec:drift}, in order to keep passivity, the proposed compensator is only able to reduce the drift when energy gaps are present. During the moments when the passivity controller is acting to reduce the delayed master velocity coming from the channel, the accumulation of drift is unavoidable. For that reason, this section aims to analyze the convergence characteristics of the compensator during the moments where it is allowed to act. 
\par 
At the moments where the compensation action is allowed, (\ref{eq:drift_added}) becomes
\begin{equation} \label{eq:cmp}
\AdE(k) \, \D\mbi{V}_{sd}(k)=\, \Dh\Tilde{\mbi{V}}_{sd}(k) + \Dh \mbi{V}_{ad} (k)\, .
\end{equation}
By defining a velocity error $\mbi{V}_{E}(k)$, (\ref{eq:cmp}) becomes
\begin{equation} \label{eq:eq}
\Dh \mbi{V}_{E}(k) \triangleq  \AdE(k) \, \D\mbi{V}_{sd}(k) - \Dh\Tilde{\mbi{V}}_{sd}(k) = \Dh \mbi{V}_{ad} (k)\, .
\end{equation}
From this definition, the error pose $\mbi{g}_E(k)$ can be defined as in (\ref{eq:int_space}) with $\Dh \mbi{V}_{E}(k)$ as its spatial velocity as follows
\begin{equation}
\label{eq:He_int}
   \mbi{g}_{E}(k)= \expE\bigl([\Dh\mbi{V}_{E}(k)]^{\wedge}\Delta T\bigr) \mbi{g}_{E}(k-1)\, .
\end{equation}
By exploring the equality between $\Dh \mbi{V}_{E}(k)$ and $\Dh \mbi{V}_{ad} (k)$ defined in (\ref{eq:eq}), the error pose from (\ref{eq:He_int}) becomes
\begin{equation}
\label{eq:He_vad}
   \mbi{g}_{E}(k)= \expE\bigl([\Dh\mbi{V}_{ad}(k)]^{\wedge}\Delta T\bigr) \mbi{g}_{E}(k-1)\, .
\end{equation}
It follows from the compensation law (\ref{eq:vad}) and the definition of the exponential function in $SE(3)$ (\ref{eq:exp_se3}) that the rotational part of (\ref{eq:He_int}) becomes
\begin{equation}
\label{eq:Re_comp}
    \mbi{R}_{E}(k)= \expO\left( -\mbi{k_R} \,\widehat{\varphi}_E(k-1) \right) \mbi{R}_{E}(k-1) \, ,
\end{equation}
which results in the following relation
\begin{equation}
\label{eq:drift_dyn}
    \varphi_E(k)=(1-\mbi{k_R})\varphi_E(k-1) \, .
\end{equation}
 Likewise, the translational part becomes
\begin{equation} \label{eq:pe_comp}
\begin{split}
    \mbi{p}_{E}(k) &= \expO\left(\widehat{\omega}_{ad}(k) \Delta T\right) \mbi{p}_E(k-1)  \\ - &\mbi{A}\left(\omega_{ad}(k) \Delta T\right)\,  \mbi{A}\left(\omega_{ad}(k) \Delta T\right)^{-\mathit{T}} \mbi{K_T} \, \mbi{p}_E(k-1)\, \, .
    \end{split}
\end{equation}
By using the identity from (\ref{eq:A_id}), (\ref{eq:pe_comp}) becomes
\begin{equation}
    \mbi{p}_{E}(k) = \expO\left(\widehat{\omega}_{ad}(k) \Delta T\right) \left( \mathbf{I} - \mbi{K_T} \right) \, \mbi{p}_E(k-1)\, \, .
\end{equation}
\par
It can be verified that a sufficient condition for convergence is 
\begin{equation}
    0 < \: \mbi{k_R}\:  < 2 \quad \wedge \quad
    0 < \, \text{eig}(\mbi{K_T}) \, <2 \, ,
\end{equation}
which ensures that
\begin{align}
    ||\varphi_E(k)||<||\varphi_E(k-1)|| \, ,\\
    ||\mbi{p}_E(k)||<||\mbi{p}_E(k-1)|| \, ,
\end{align}
as long as the trace of the accumulated rotational error $\mbi{R}_E$ is not equal to one, when the compensator is allowed to act after the drift has been accumulated by the passivity controller. The above presented compensation law makes sure that the magnitude of the drift is decreased from one time step to the next, even if the compensator is only allowed to act during a short period of time. 
\par
It is also interesting to note that, if the gain matrices are chosen to be identity matrices, the accumulated drift becomes zero within one time step. This can be seen as the multi-DoF extension of the compensators proposed in \cite{artigas10} and \cite{chawda14}. However, in case the force peaks described in \cite{chawda14} are undesirable, other values within the convergent range could be chosen. In that case, the proposed compensator can be considered as an extension of the one previously presented by the authors in \cite{coelho18}.

\section{Validation Results}
\label{sec:val}
\subsection{Overview}
This section provides experimental results performed using two 3-DoF devices (Section~\ref{sec:exp}), as well as numerical simulation results of teleoperation of the model of the Suspended Aerial Manipulator from Fig.~\ref{fig:sam} (Section~\ref{sec:sim}). In order to show the efficacy of the proposed compensator when both concatenated and coupled passivity controller approaches are applied, Section~\ref{sec:exp} focuses on the implementation of the former while Section~\ref{sec:sim} shows results for the latter.


\subsection{Experimental Evaluation}
\label{sec:exp}
\begin{figure}[t]
    \centering
    \includegraphics[height=0.4\linewidth]{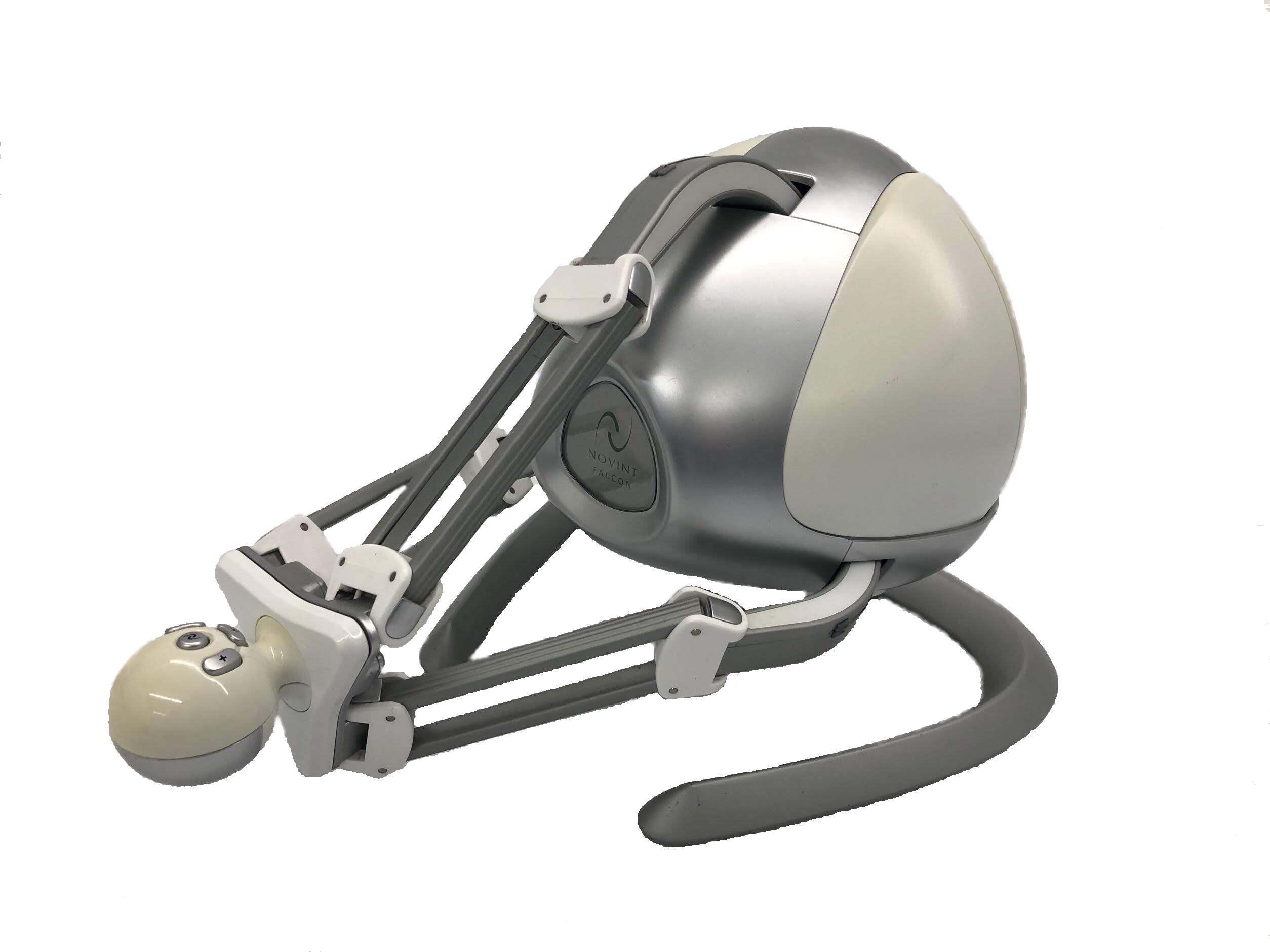}
    \caption{Haptic device used in validation experiments.}
    \label{fig:falcon}
\end{figure}

In order to validate the proposed compensator on multi-DoF devices, telemanipulation experiments were performed using two 3-DoF translational \textit{Novint Falcon} haptic devices (see Fig.~\ref{fig:falcon}). Firstly, the concatenated PO-PC alone was applied to passivate the communication channel, set to artificially add 200 ms round-trip time delays ($T_{rt}$). Subsequently, the translational part of the proposed compensation law (\ref{eq:vad}) with $\mbi{\omega}_{ad}=0$ was applied. 
\par
\begin{figure*}[tb]
\begin{subfigure}{0.33\linewidth}
\centering
\includegraphics[trim={0cm 0cm 0cm 0cm},clip,width=0.88\linewidth]{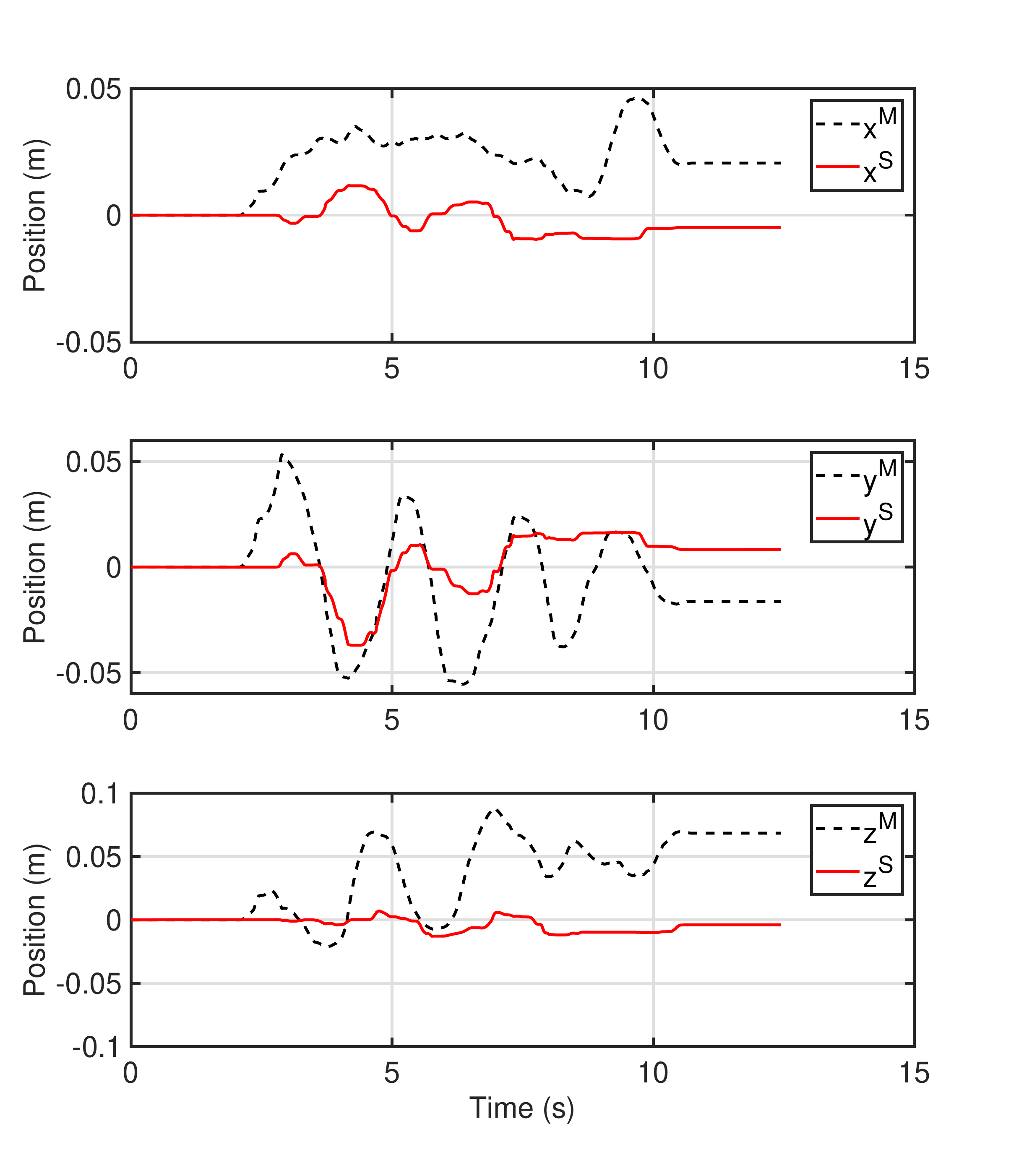}
\caption{}
\label{fig:pospc200}
\end{subfigure}%
~
\begin{subfigure}{0.33\linewidth}
\includegraphics[trim={0cm 0cm 0cm 0cm},clip,width=0.88\linewidth]{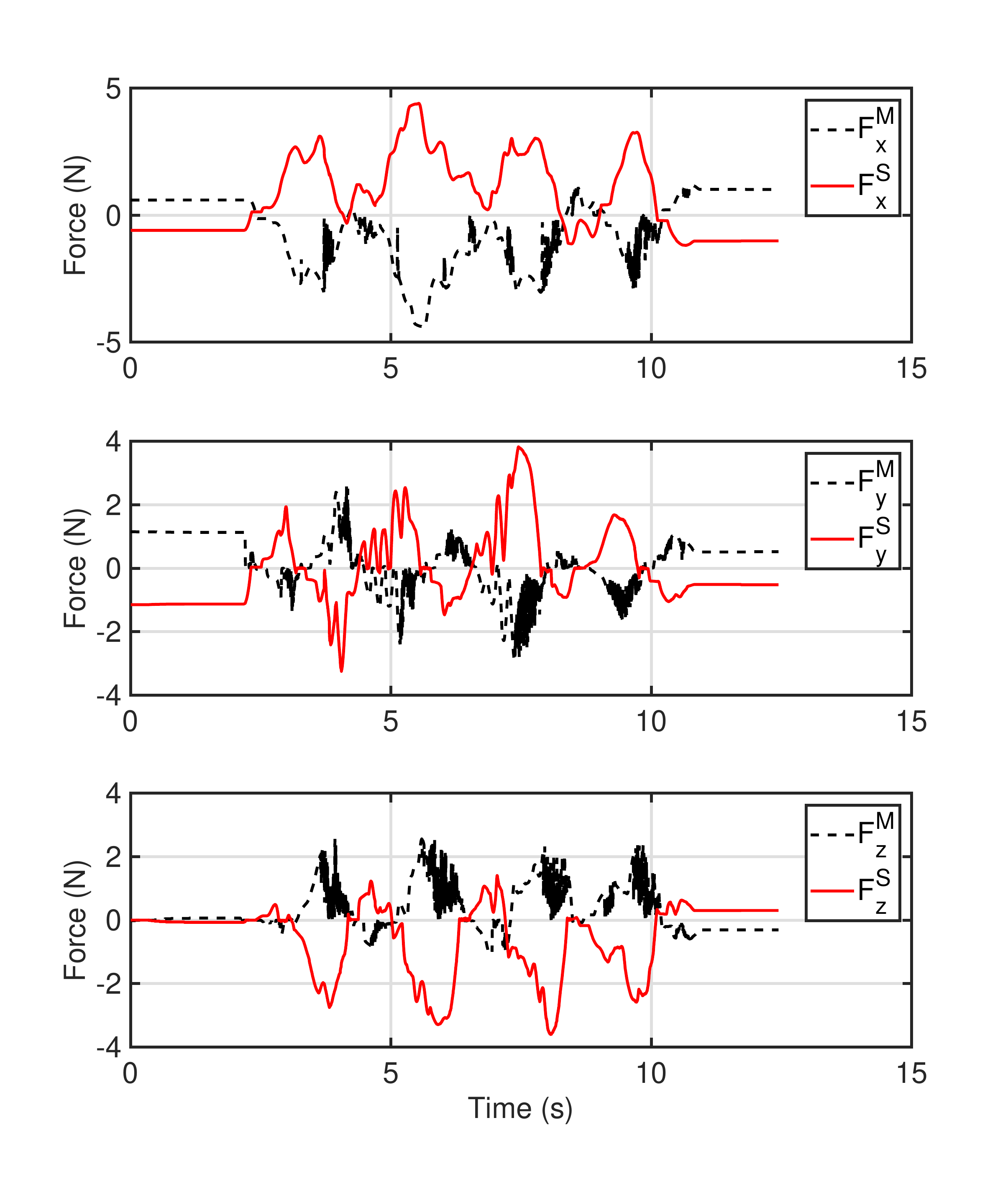}
\caption{}
\label{fig:fpc200}
\end{subfigure}
\begin{subfigure}{0.33\linewidth}
\centering
 \includegraphics[trim={0cm 0cm 0cm 0cm},clip,width=0.88\linewidth]{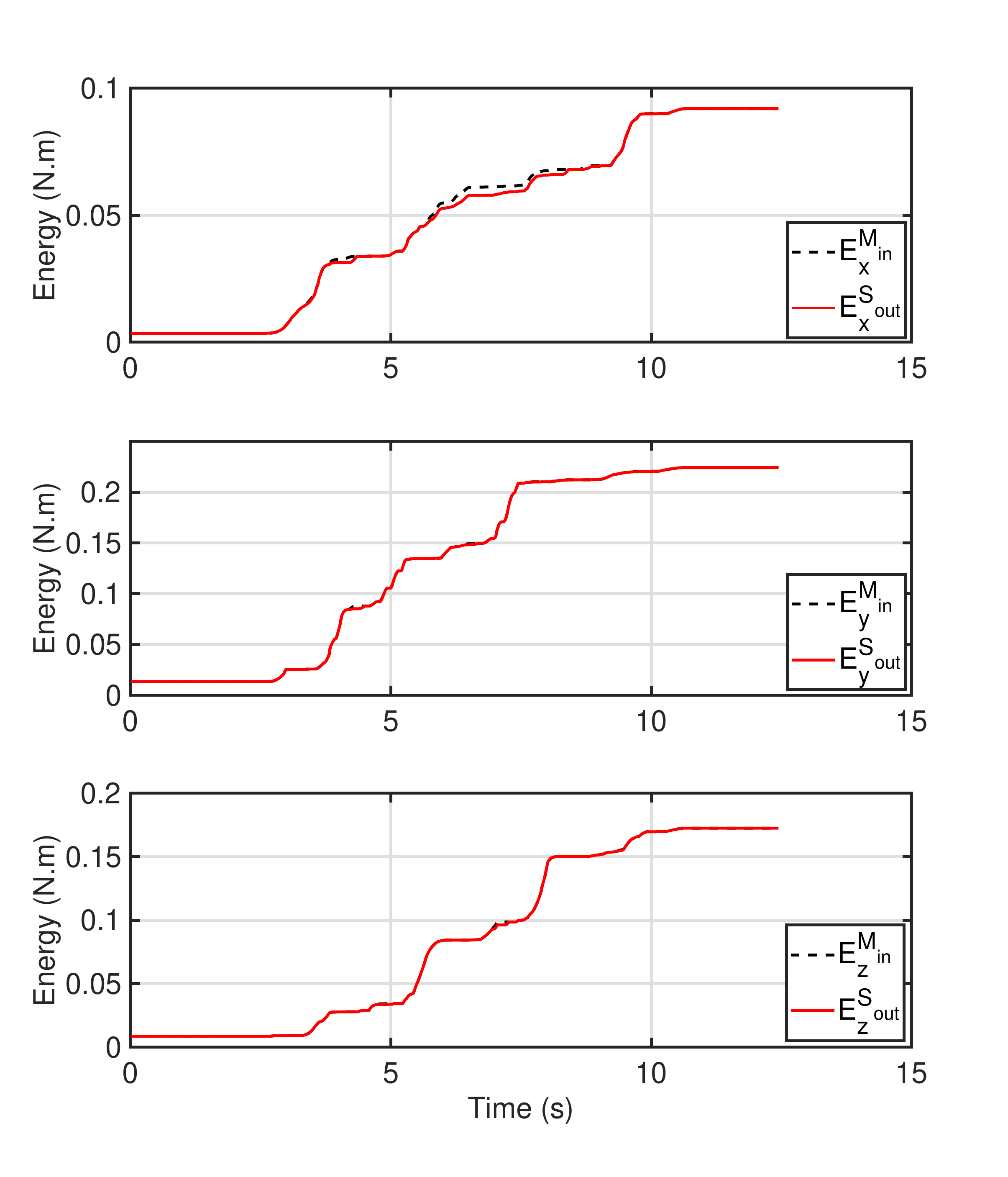}
 \caption{}
\label{fig:EMinpc200}
\end{subfigure}
\caption{No drift compensator -- $T_{rt}=200$ ms. (a) master and slave positions, (b) master and slave forces, (c) master-in and slave-out energies.}
\end{figure*}

Figs.~\ref{fig:pospc200} and \ref{fig:fpc200} show position and control forces, respectively, of the master and slave devices for the non-compensated case. Fig.~\ref{fig:EMinpc200} shows the master-input and slave-output energies observed on the slave side. It can be seen that, in order to ensure passivity of the channel, not only the impedance-type PC intermittently reduced the force values (Fig.~\ref{fig:fpc200}), but also the admittance-type PC removed part of the velocity coming from the master, generating significant drift (see Fig.~\ref{fig:pospc200}).

\par
\begin{figure*}[tb]
\centering
\begin{subfigure}[thpb]{0.33\linewidth}
\includegraphics[trim={0cm 0cm 0cm 0cm},clip,width=0.88\linewidth]{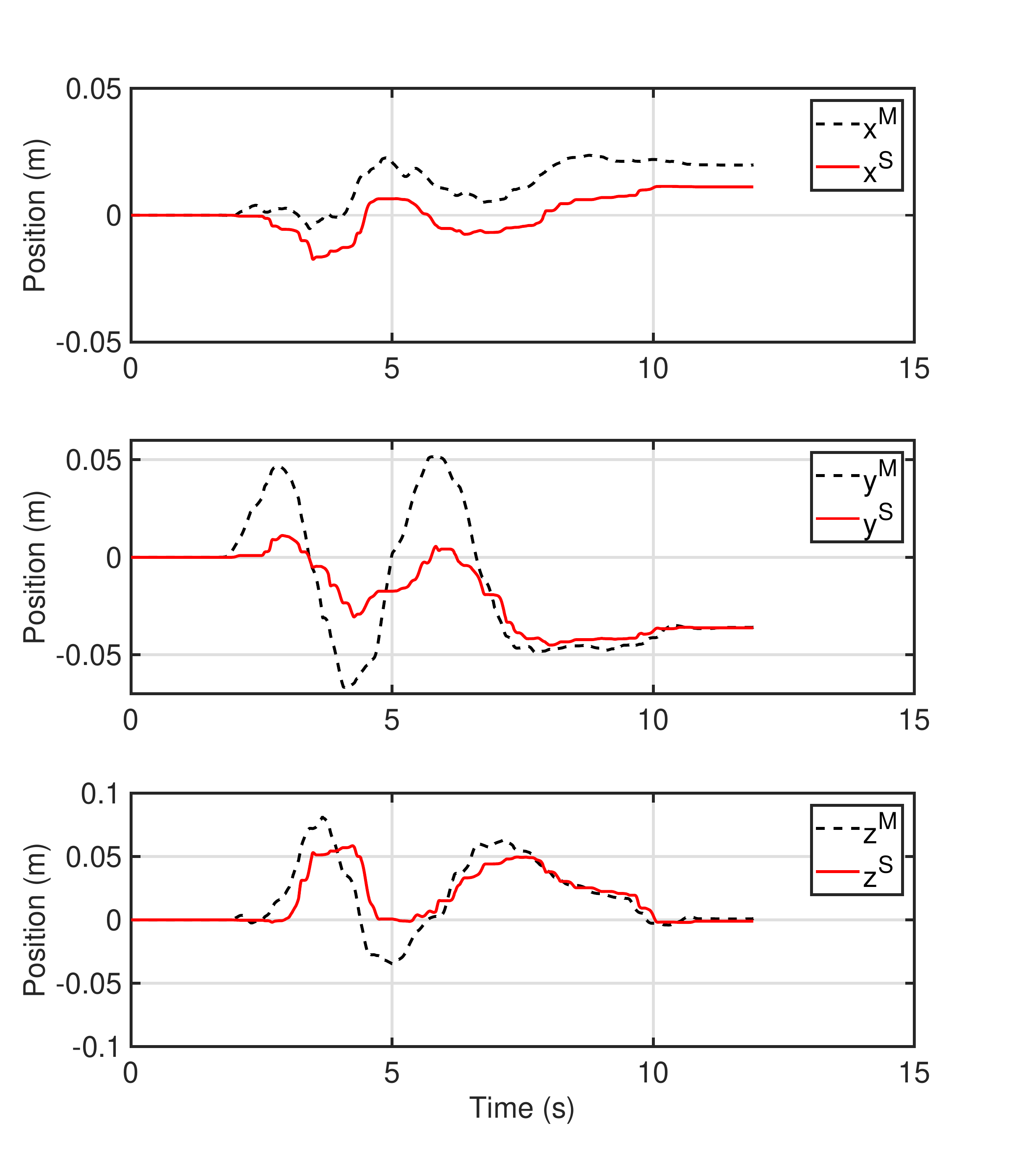}
\caption{}
\label{fig:posdc200}
\end{subfigure}%
~
\begin{subfigure}[thpb]{0.33\linewidth}
 \includegraphics[trim={0cm 0cm 0cm 0cm},clip,width=0.88\linewidth]{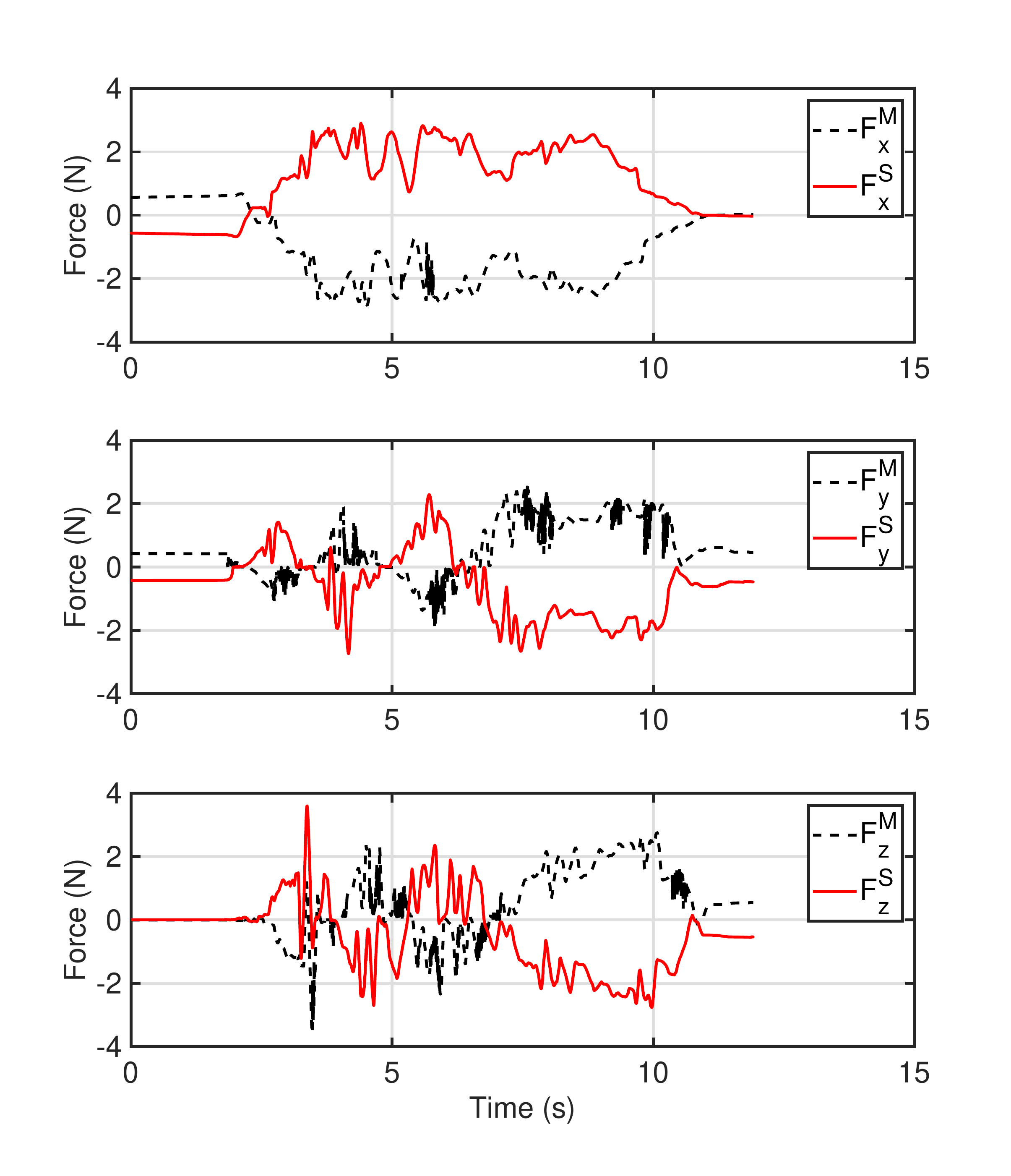}
 \caption{}
\label{fig:fdc200}
\end{subfigure}%
~
\begin{subfigure}[thpb]{0.33\linewidth}
\centering
\includegraphics[trim={0cm 0cm 0cm 0cm},clip,width=0.88\linewidth]{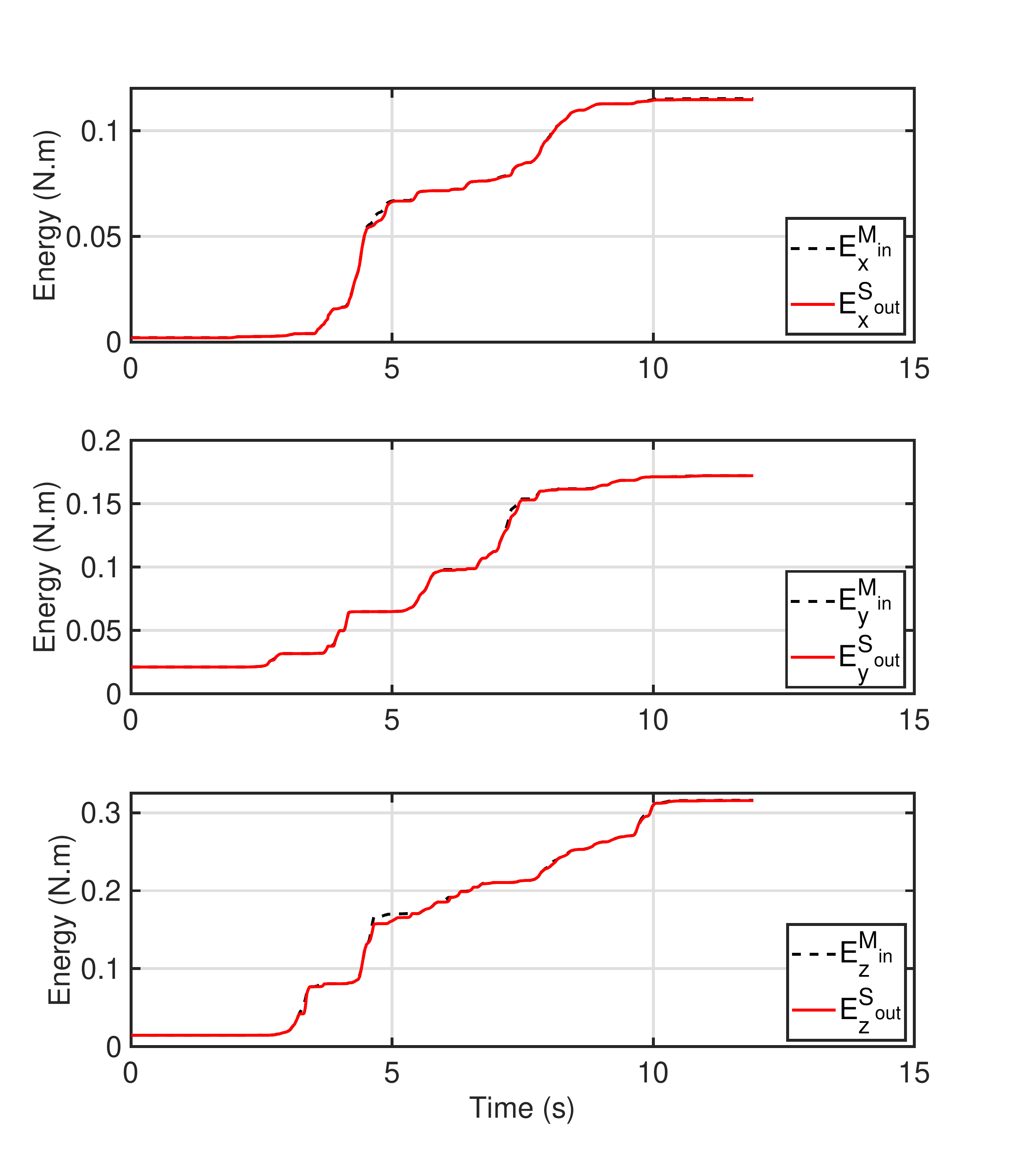}
\caption{}
\label{fig:EMindc200}
\end{subfigure}
\caption{Drift compensator on -- $T_{rt}=200$ ms. (a) master and slave positions, (b) master and slave forces, (c) master-in and slave-out energies.}
\end{figure*}

Figs.~\ref{fig:posdc200}--\ref{fig:EMindc200} show position, force, and energy values, respectively, for the case when the proposed drift compensator was applied. It can be seen that the compensator was able to completely remove the drift in the y- and z-axes (Fig.~\ref{fig:posdc200}). However, an offset can still be observed in the x-axis. This is due to the fact that not enough passivity gaps appeared in order to compensate for the drift in a passive way. The occurrence of passivity gaps depends on the system dynamics, the task being performed and the delay of the channel.  
\subsection{Numerical Simulation}
\label{sec:sim}
In order to show the applicability of the drift compensation law, the proposed compensator was applied to the Cartesian pose of the end-effector of a simulated Suspended Aerial Manipulator (Fig.~\ref{fig:sam}) when round-trip communication delays of 700 ms were present. In the results presented in this section, the coupled PO-PC implementation was used (see Section~\ref{sec:pc}). \par
Despite being a redundant manipulator, a decoupling control law was applied so that the nullspace dynamics would not affected the Cartesian-space task (see Section~\ref{sec:dyn}).
\par
\begin{figure*}[thb]
\centering
\begin{subfigure}{0.33\linewidth}
\includegraphics[trim={0cm 0cm 0cm 0cm},clip,width=0.88\linewidth]{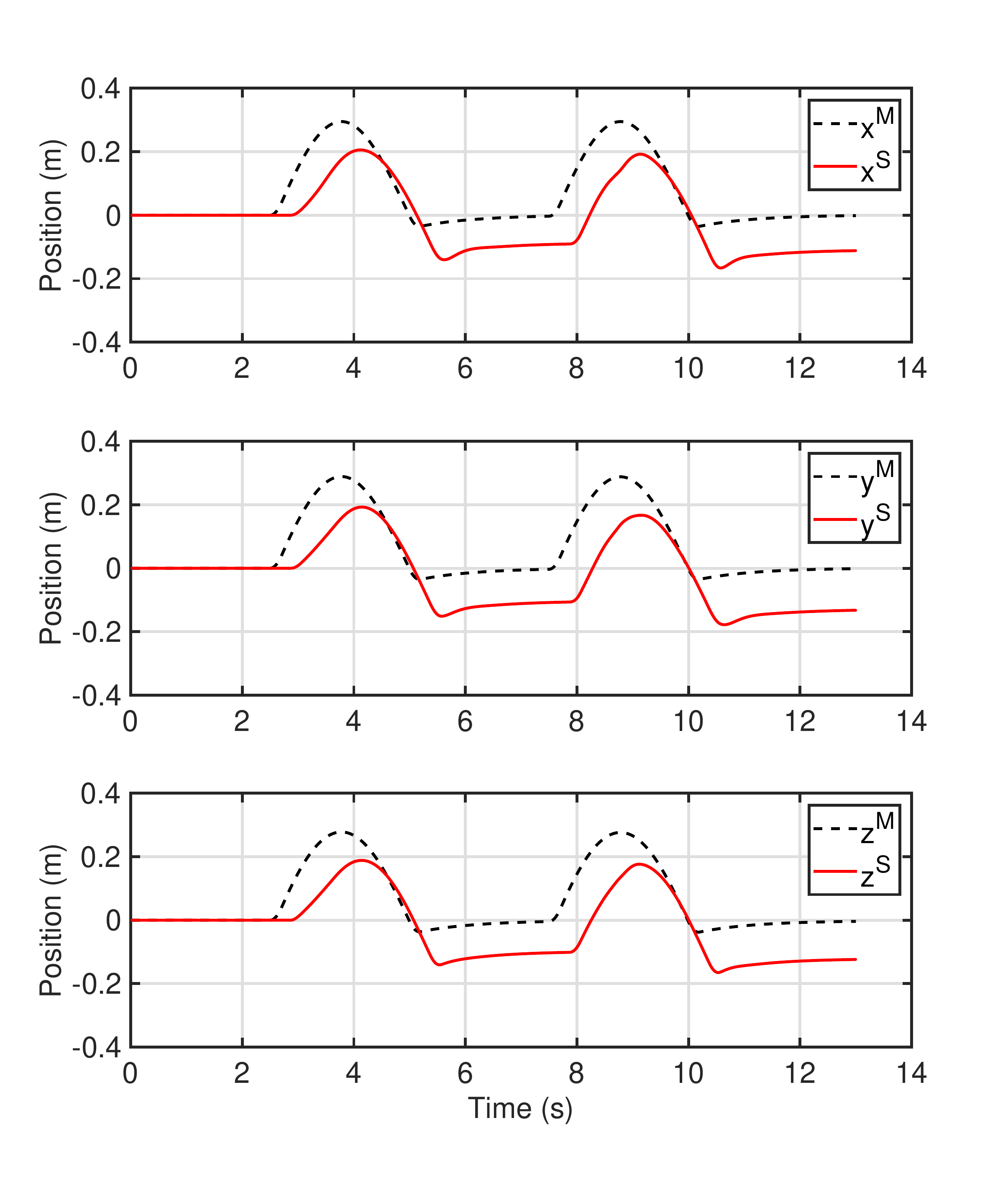}
\caption{}
\label{fig:pospc700}
\end{subfigure}%
~
\begin{subfigure}{0.33\linewidth}
\includegraphics[trim={0cm 0cm 0cm 0cm},clip,width=0.88\linewidth]{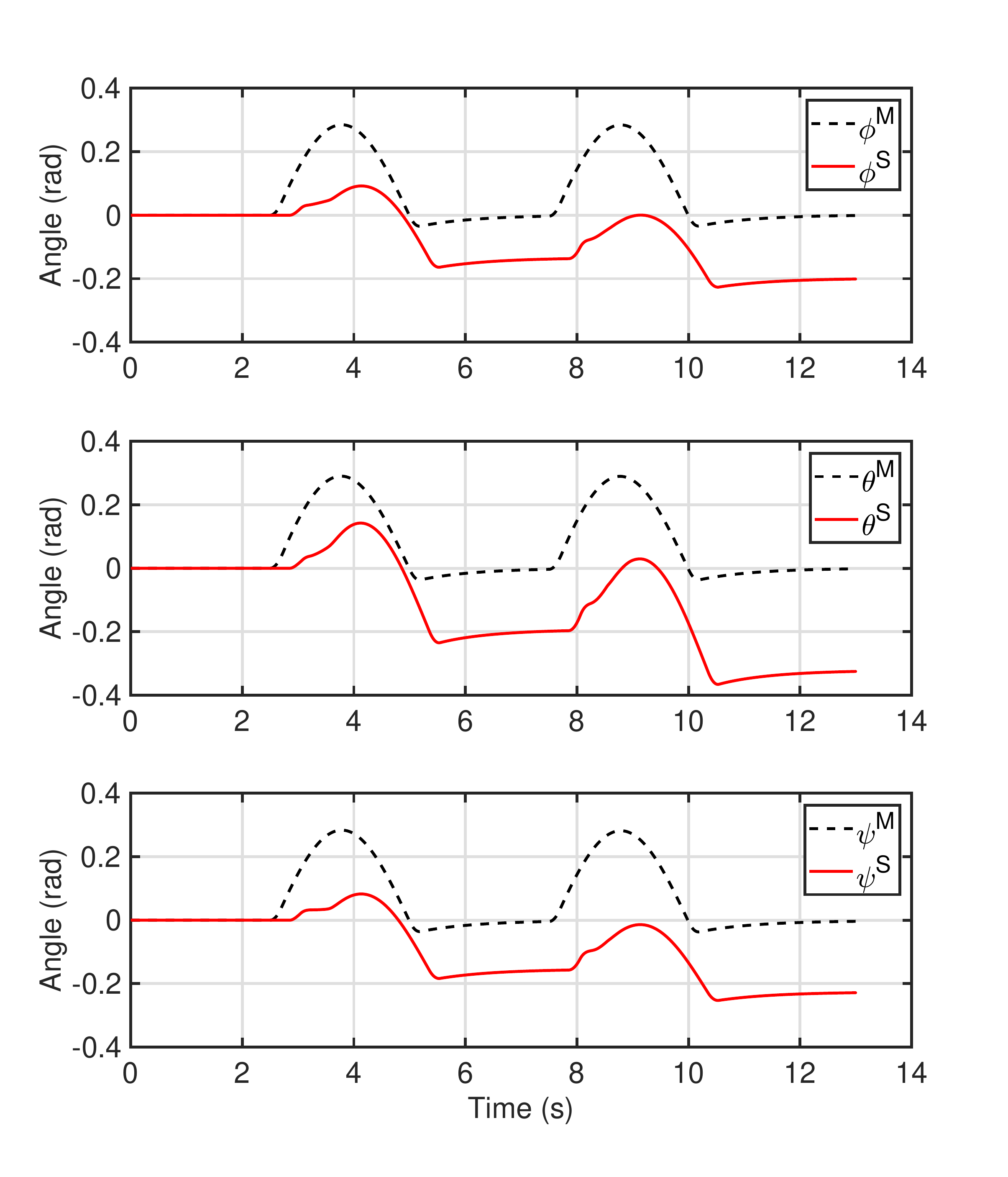}
\caption{}
\label{fig:RPYpc700}
\end{subfigure}%
~
\begin{subfigure}{0.33\linewidth}
\includegraphics[trim={0cm 0cm 0cm 0cm},clip,width=0.88\linewidth]{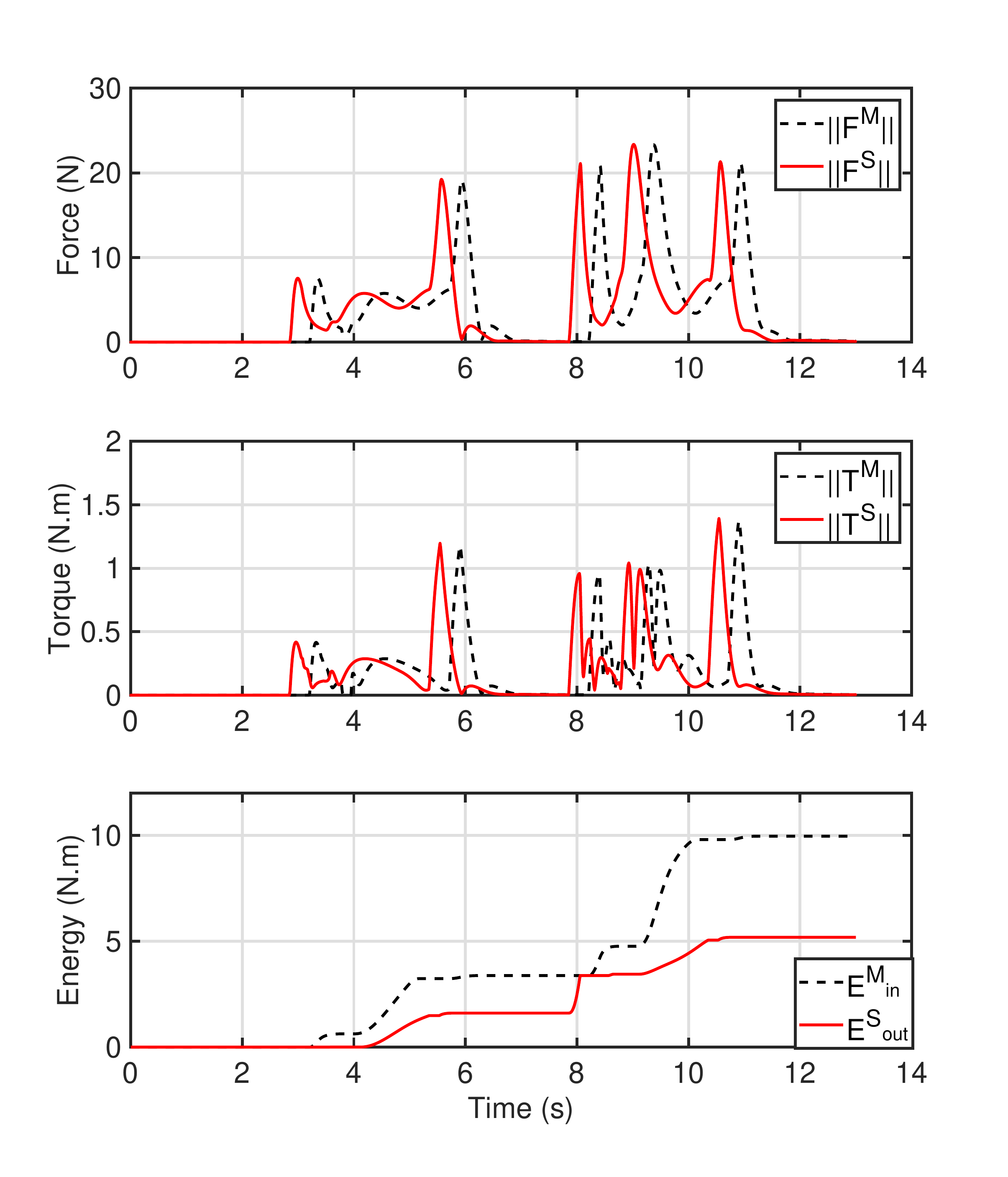}
\caption{}
\label{fig:F_En_pc700}
\end{subfigure}
\caption{No drift compensator  -- $T_{rt}=700$ ms. (a) master and slave positions, (b) master and slave orientation, (c) Euclidean norm of master and slave body-frame Cartesian forces (top) and torques (middle),  master-in and slave-out energies (bottom).}
\end{figure*}

Figs~\ref{fig:pospc700}--\ref{fig:F_En_pc700} depict the end-effector pose, the Euclidean norm of the tool-frame Cartesian forces and torques, and slave-side energy signals computed when applying coupled TDPA without drift compensation. It is important to note that Roll-Pitch-Yaw (RPY) angles were used in order to facilitate the understanding of the orientation plots. From Figs.~\ref{fig:pospc700} and \ref{fig:RPYpc700}, significant drift caused by the admittance-type PC can be observed in both position and orientation values. It can be noted that the deviation between master and slave poses increased significantly when nonzero references were given.
\par
\begin{figure*}[t]
\centering
\begin{subfigure}{0.33\linewidth}
\includegraphics[trim={0cm 0cm 0cm 0cm},clip,width=0.88\linewidth]{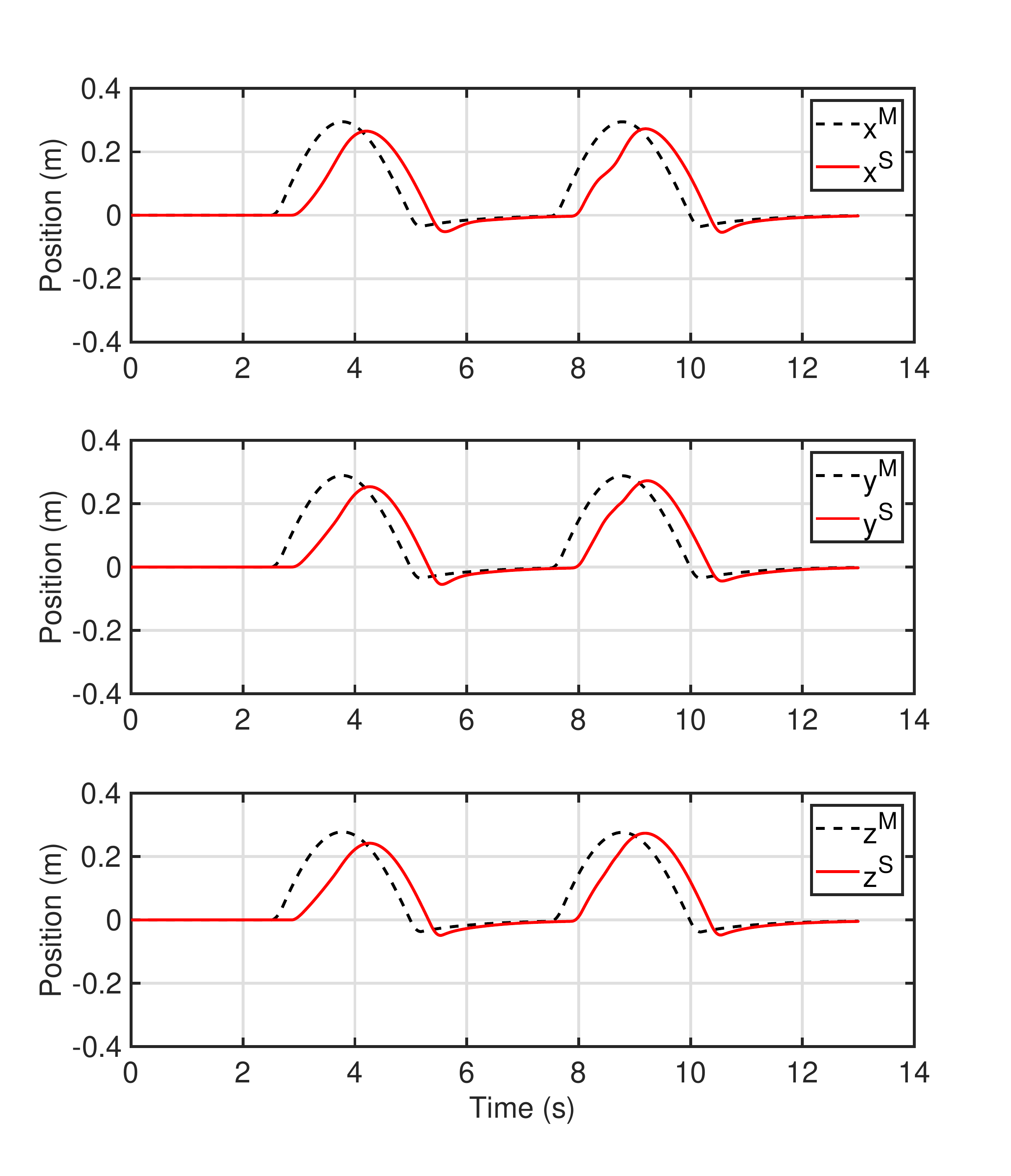}
\caption{}
\label{fig:posdc700}
\end{subfigure}%
~
\begin{subfigure}{0.33\linewidth}
\includegraphics[trim={0cm 0cm 0cm 0cm},clip,width=0.88\linewidth]{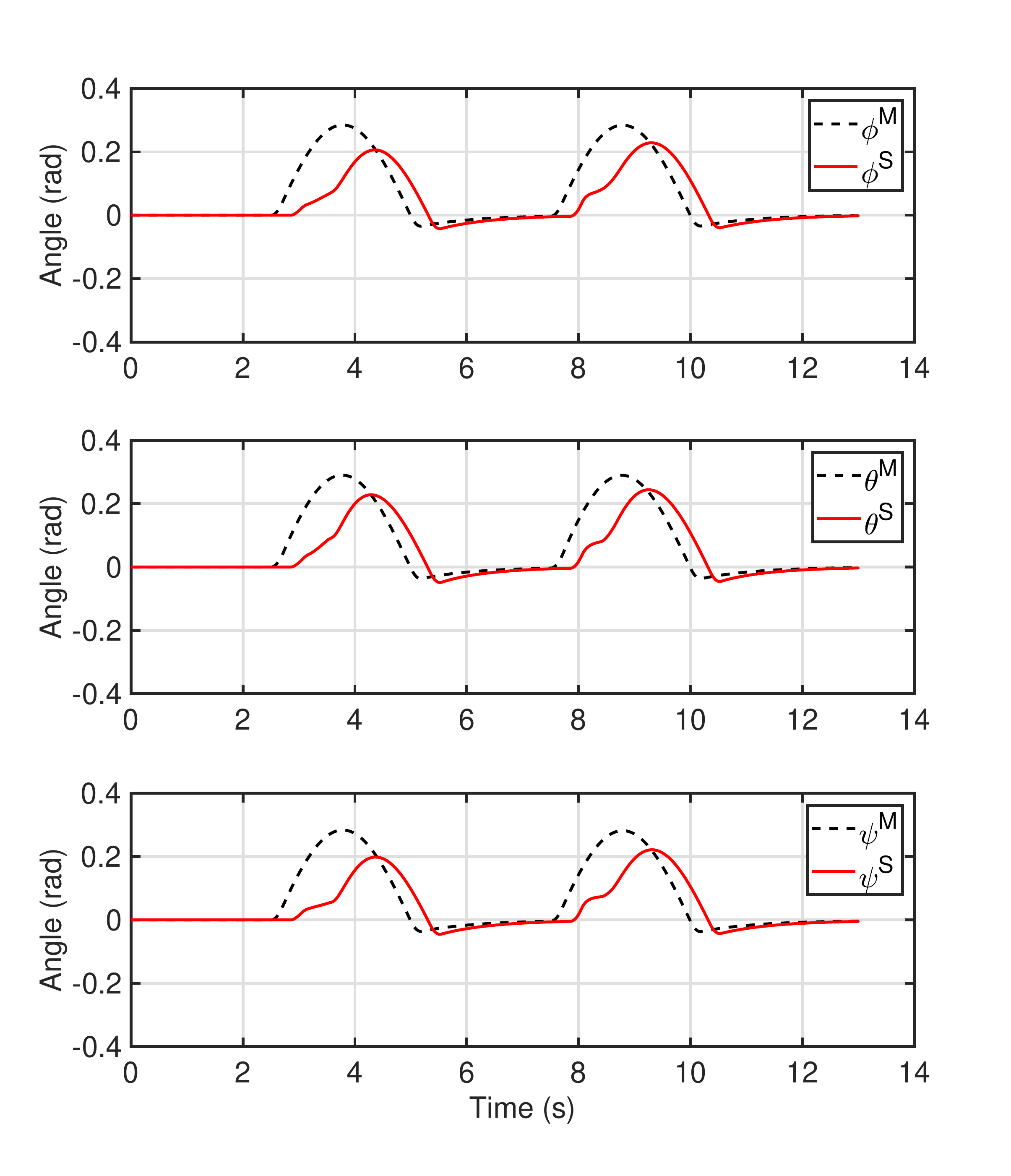}
\caption{}
\label{fig:RPYdc700}
\end{subfigure}%
~
\begin{subfigure}{0.33\linewidth}
\includegraphics[trim={0cm 0cm 0cm 0cm},clip,width=0.88\linewidth]{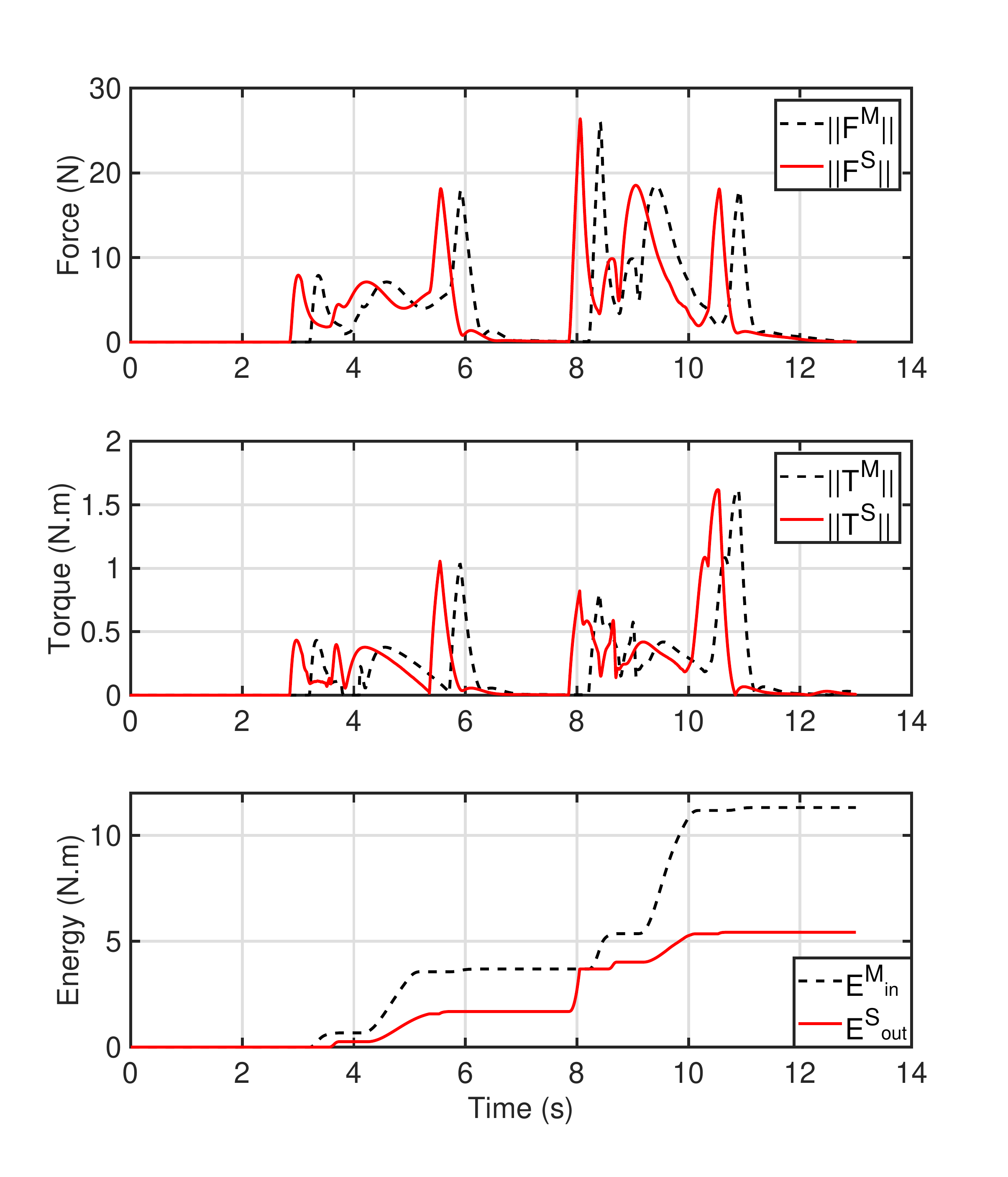}
\caption{}
\label{fig:F_En_dc700}
\end{subfigure}
\caption{Drift compensator on -- $T_{rt}=700$ ms. (a) master and slave positions, (b) master and slave orientation, (c) Euclidean norm of master and slave body-frame Cartesian forces (top) and torques (middle),  master-in and slave-out energies (bottom).}
\end{figure*}

When drift compensation was added (Figs.~\ref{fig:posdc700}--\ref{fig:F_En_dc700}), it can be noted that the drift converged to zero in both position and orientation (Figs.~\ref{fig:posdc700} and \ref{fig:RPYdc700}). It is also important to remark that the norm of the control forces and torques (Fig.~\ref{fig:F_En_dc700}) were not increased significantly when the compensator was applied. Adding to that, it can be seen that the compensator was able to reduce the drift without compromising the passivity of the system (see the energy plot in Fig.~\ref{fig:F_En_dc700}). 




\section{Conclusion}
\label{sec:conc}
This paper presented an extension of the previously proposed TDPA-based drift compensators to multi-DoF Cartesian-Space teleoperation. A convergence analysis has also been provided. It has been shown that, if the gain is set within certain bounds, the proposed approach is able to reduce the drift caused by the passivity controller in case it is able to do so without violating the passivity condition. That analysis also provided an insight about the cause of force spikes, which are generated when the drift is set to converge within one time step. In addition, hardware experiments and numerical simulation results demonstrated the applicability of the proposed compensator to time-delayed bilateral teleoperation of multi-DoF devices, when using both concatenated and coupled PO-PC implementations. Future work will involve applying teleoperation methods to the nullspace of redundant manipulators.

\bibliographystyle{IEEEtran}
\bibliography{root}
\end{document}